\documentclass[twosides]{article}
\usepackage{amsmath}
\usepackage{graphicx}
\usepackage{caption}
\usepackage{booktabs,tabularx}
\usepackage[numbers,sort&compress]{natbib}
\usepackage{multirow}
\usepackage[section]{placeins}

\usepackage{PRIMEarxiv}

\usepackage[utf8]{inputenc} 
\usepackage[T1]{fontenc}    
\usepackage{hyperref}       
\usepackage{url}            
\usepackage{booktabs}       
\usepackage{amsfonts}       
\usepackage{nicefrac}       
\usepackage{microtype}      
\usepackage{lipsum}
\usepackage{fancyhdr}       
\usepackage{graphicx}       
\graphicspath{{Images/}}     

\title{Co-design is powerful and not free
\thanks{Corresponding author: \texttt{stao@tju.edu.cn}} }

\author{
  Yi Zhang \\
  Key Laboratory of Mechanism Theory \\and Equipment Design of Ministry of Education\\
  Tianjin University \\
  Tianjin, China\\
  \texttt{zhangyi\_@tju.edu.cn} \\
   \And
  Yue Xie \\
  Department of Computer Science\\
  Loughborough University\\
  UK\\
  \texttt{y.xie4@lboro.ac.uk} \\
  \AND
  Tao Sun \\
  Key Laboratory of Mechanism Theory \\and Equipment Design of Ministry of Education\\
  Tianjin University \\
  Tianjin, China\\
  \texttt{stao@tju.edu.cn} \\
  \And
  Fumiya Iida \\
  Bio-Inspired Robotics Lab \\
  University of Cambridge \\
  Cambridge, UK\\
  \texttt{fi224@cam.ac.uk}\\
    }

\begin{document}
\maketitle

\begin{abstract}
Robotic performance emerges from the coupling of body and controller, yet it remains unclear when morphology–control co-design is necessary. We present a unified framework that embeds morphology and control parameters within a single neural network, enabling end-to-end joint optimization. Through case studies in static-obstacle-constrained reaching, we evaluate trajectory error, success rate, and collision probability. The results show that co-design provides clear benefits when morphology is poorly matched to the task, such as near obstacles or workspace boundaries, where structural adaptation simplifies control. Conversely, when the baseline morphology already affords sufficient capability, control-only optimization often matches or exceeds co-design. By clarifying when control is enough and when it is not, this work advances the understanding of embodied intelligence and offers practical guidance for embodiment-aware robot design.
\end{abstract}

\section{Introduction}

Robots must coordinate both their bodies and controllers to operate effectively in complex environments. Yet a fundamental question remains unresolved: when is controller optimization alone sufficient, and when does morphology–control co-design become necessary? 
Traditional approaches address this challenge by fixing morphology and optimizing control, a strategy that has yielded powerful controllers but is inherently limited when the body is poorly matched to task demands. Co-design promises to overcome this mismatch, but existing studies often evaluate it within locomotion tasks or rely on two-layer optimization, where morphology and control are tuned in separate loops. These settings obscure the boundary between the strengths of control and those of co-design.  

Here we revisit this boundary by embedding morphology and control parameters within a single neural network for end-to-end joint optimization. To ensure fair comparison, we introduce two complementary protocols: one that fixes network width and one that fixes parameter count. This design allows us to distinguish whether performance differences arise from control capacity or from the integration of morphology and control.

Our analysis highlights both the strengths and limitations of co-design. In some cases, adapting morphology simplifies the control problem and improves robustness. In others, controller optimization alone suffices, and additional morphological search only slows convergence. By distinguishing when control is enough and when it is not, this study contributes a clearer understanding of the boundary conditions that govern the utility of co-design in manipulation tasks.

This article is organized as follows. First, we review prior work on control-only optimization and morphology–control co-design, highlighting both their progress and limitations. Next, we describe the experimental setup, including the simulation platform, task definition, and evaluation metrics. The core of the paper presents case studies that contrast co-design and control-only approaches under static-obstacle-constrained reaching tasks. Finally, we discuss the insights gained from these analyses, emphasizing the boundary conditions that clarify when control is enough and when it is not.


\section{Related Work}

Robotic morphology and control are deeply coupled, especially in environments with dense obstacles or narrow passages~\cite{PfeiferBongard2006}. Traditional design follows a sequential paradigm: the body is fixed first, and the controller is optimized later. This has led to a wide range of effective controllers. Classical PID control remains popular for its simplicity and ease of tuning~\cite{ZieglerNichols1942,AstromHagglund1995}, but its performance degrades in highly nonlinear or cluttered settings. Model Predictive Control (MPC) incorporates dynamics and constraints explicitly, enabling precise tracking and obstacle avoidance~\cite{GarciaMorari1989}, yet it is computationally demanding and cannot overcome limitations imposed by the body’s reach or geometry. More recently, reinforcement learning has shown strong adaptability in standardized benchmarks~\cite{Brockman2016Gym,Tassa2018DMC,Schulman2017PPO}, but these tasks typically treat morphology as fixed, leaving the learned policies brittle when structural mismatches arise.  

The common assumption behind these control-only approaches is that the morphology is already well matched to the task. When this is not the case, when the workspace is too small, the kinematics ill-suited, or the payload insufficient—no controller can fully compensate. Failures manifest as unreachable targets, excessive collisions, or unstable motions. Moreover, traditional static design metrics such as workspace coverage or force transmission efficiency capture only limited aspects of performance and generalize poorly to novel or cluttered environments.  

These limitations highlight that part of skill and intelligence resides in the body itself. Adjusting morphology can open collision-free corridors, improve manipulability, or reduce control effort. Consequently, recent work has shifted toward \textbf{morphology--control co-design}, where structure and policy are optimized together to achieve more integrated and adaptive performance.

\subsection{Control-only Optimization Approaches}

Classical PID controllers remain widely used for their simplicity and robustness, with procedures such as Ziegler–Nichols tuning ensuring broad applicability~\cite{ZieglerNichols1942,AstromHagglund1995}. Model Predictive Control (MPC) further advanced trajectory tracking and interaction tasks by explicitly handling dynamics and state constraints~\cite{GarciaMorari1989,RawlingsMayneDiehl2017}. More recently, reinforcement learning (RL) has accelerated controller design through standardized benchmarks such as OpenAI Gym and the DeepMind Control Suite~\cite{Brockman2016Gym,Tassa2018DMC}, which catalyzed rapid progress in policy learning. However, these platforms define tasks only at the control level, explicitly decoupled from morphology.  

When tasks are structurally constrained, for example, reaching through clutter, wrapping around obstacles, or contacting within tight clearances, control-only approaches face intrinsic limits. Even optimal PID or MPC policies cannot compensate for morphological mismatches in workspace, kinematics, or payload. MPC enforces constraints effectively, but the feasible set remains bounded by the given body. Empirical studies confirm strong MPC performance in manipulation~\cite{Luo2024TubeMPC}, but also highlight latency trade-offs without structural adaptation. From an embodied intelligence perspective, morphology and control co-shape capability~\cite{PfeiferBongard2006}, leading to diminishing returns when optimizing control in isolation.  

Taken together, PID and MPC are indispensable baselines and often state of the art on fixed platforms. Yet in obstacle-rich manipulation, substantial headroom remains that likely requires \textbf{morphology--control co-design} rather than control-only tuning.

\subsection{Studies on Morphology--Control Co-Design}

With the growing recognition of control-only limitations, morphology--control co-design has emerged, optimizing the robot's structure and controller jointly. Sims' pioneering work in the 1990s co-evolved virtual creatures, revealing the strong coupling between form and behavior~\cite{Sims1994EvolvingCreatures}. More recently, Schaff et al.~\cite{Schaff2019ConstructControl} proposed a DRL-based method that learns to build and control agents simultaneously, producing novel locomotion designs that surpassed fixed baselines. Wang et al.~\cite{Wang2019NGE} introduced Neural Graph Evolution, which evolves robots as graphs and leverages GNN controllers to adapt quickly, achieving faster search and discovering intuitive designs such as finned fish and cheetah-like morphologies. Both studies tackled the challenge of expensive inner-loop training by enabling skill reuse across morphologies.  

Beyond locomotion, Pathak et al.~\cite{Pathak2019SelfAssembling} studied modular self-assembling morphologies, where limb modules connect and disconnect under a single policy, showing stronger generalization than static robots. Cheney et al.~\cite{Cheney2014SoftRobots} treated shape change itself as an action in soft robots, further blurring the boundary between design and control and reinforcing embodied intelligence.  

Most co-design research remains mobility-centric. RoboGrammar~\cite{RoboGrammar2020} used graph grammars to generate terrain-specific legged robots, yielding non-intuitive yet effective morphologies. For manipulation, systematic co-design is still nascent. Early work includes Yi et al.~\cite{Yi2025SoftGripper}, who co-designed material stiffness and grasp poses of soft grippers, significantly outperforming manual baselines in both simulation and physical tests.

\section{Case-Oriented Setup}

\subsection{Problem and Method Overview}

We cast morphology--control co-design as joint optimization over morphology parameters $\theta^m$ and control parameters $\theta^c$ \emph{within a single neural policy}, in contrast to bi-level schemes. In the co-design condition, $(\theta^m,\theta^c)$ are encoded in one decision vector and optimized synchronously; the policy is conditioned on morphology via inputs $[s_t, x^*, \theta^m]$. In the control-only condition, morphology is fixed to a baseline $\bar{\theta}^m$ and omitted from the policy input, optimizing only $\theta^c$.

Let $s_t$ denote the robot state at time $t$, $u_t$ the control input, $p_t(\theta^m)$ the end-effector position under morphology $\theta^m$, and $x^*$ the target position. The MuJoCo-based simulator induces the dynamics
\[
s_{t+1} \;=\; f_{\theta^m}(s_t, u_t), 
\qquad 
u_t \;=\; \pi_{\theta^c}\!\big([\,s_t,\, x^*,\, \theta^m\,]\big),
\]
where the $\theta^m$ term in the policy input is used only in the co-design setting. Task performance is quantified by a finite-horizon trajectory loss
\[
L_{\text{task}}(\theta^m,\theta^c) 
\;=\; \frac{1}{T}\sum_{t=1}^{T} \big\| p_t(\theta^m) - x^*\big\|_2^2,
\]
augmented by a weighted collision penalty $C_{\text{coll}}(\theta^m,\theta^c)$ that accumulates signed-distance violations along the trajectory. The overall objective is
\[
\min_{\theta^m,\theta^c} \; L(\theta^m,\theta^c) 
\;=\; L_{\text{task}}(\theta^m,\theta^c) \;+\; \lambda_{\text{coll}}\, C_{\text{coll}}(\theta^m,\theta^c),
\]
subject to morphology feasibility constraints. Each link length is bounded as $\ell_i \in [\ell_i^{\min},\, \ell_i^{\max}]$. Denoting the feasible set by $\mathcal{M}$, feasibility is enforced by projection $\theta^m \leftarrow \Pi_{\mathcal{M}}(\theta^m)$ when required.

This unified objective provides a common ground for comparing optimization strategies under identical metrics and constraints: in the \emph{control-only} condition we solve $\min_{\theta^c} L(\bar{\theta}^m,\theta^c)$ with fixed baseline morphology $\bar{\theta}^m$, whereas in the \emph{co-design} condition we solve the full problem $\min_{\theta^m,\theta^c} L(\theta^m,\theta^c)$. Operationally, the pipeline forms a closed loop among simulator, controller, and optimizer: rollouts generate trajectories and losses that drive updates to both $\theta^m$ and $\theta^c$, which are then fed back for the next iteration.

\begin{figure}[htbp]
    \centering
    \includegraphics[width=0.8\linewidth]{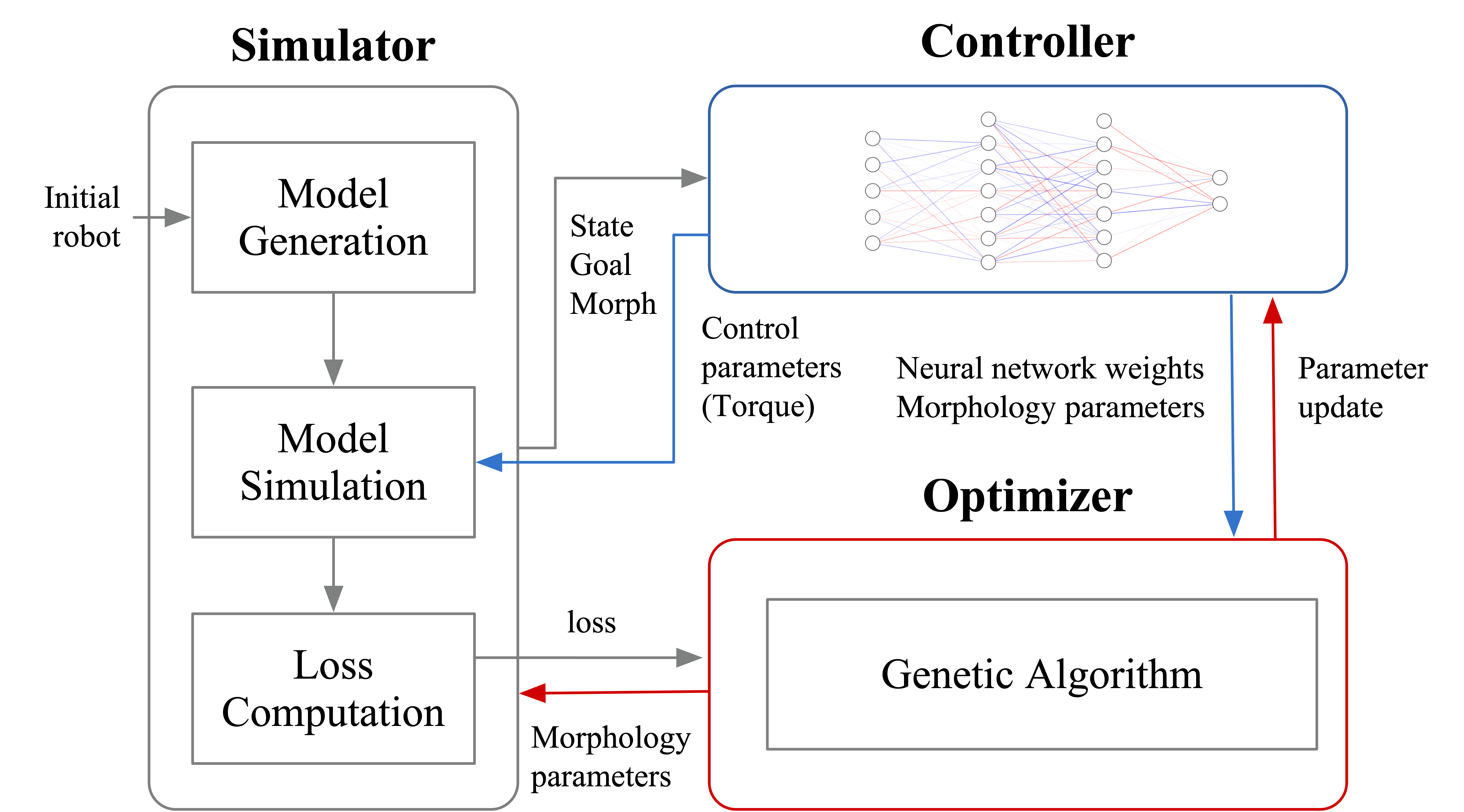}
    \caption{System Architecture Overview}
    \label{fig:figure1}
\end{figure}

\subsection{Control-only vs. Co-design}
In the Control-only setting, morphology parameters $\theta^m$ are fixed, and only the control parameters $\theta^c$ are optimized. This isolates the role of the controller but may result in limited adaptability when the morphology is mismatched with the task or environment. In this case, the neural network relies solely on the motion states to output control parameters, with morphology not included as an input. The network's input dimensionality is thus determined by the state variables, and the network is trained to optimize control actions based on this fixed morphology.

In contrast, the Co-design setting simultaneously optimizes both $\theta^m$ and $\theta^c$, allowing the controller and morphology to adapt to each other and to the task constraints. In this setting, morphology parameters are concatenated with the state variables into a single input vector, which is then fed into the neural network. This approach enables joint optimization of both control policies and morphology, with the neural network learning to adapt both the controller and the robot’s structure to improve task performance. 

The dimensionalities of the input and control parameters for both settings are as follows: in the Control-only experiment, the input dimensionality is 7 (state variables), and the control dimensionality is 2, with a total of 1378 parameters for the controller. In the Co-design experiment, the morphology parameters are included as additional inputs, increasing the total input dimensionality to 9. The network now optimizes both control and morphology parameters.

\subsection{Fairness Protocols}

We adopt two complementary protocols. Our experiments are conducted based on the first protocol.

\paragraph{Equal Network Width.} This is our primary setting. Both methods use the same hidden layer size (e.g., same number of neurons), meaning Co-Design’s controller has a slightly higher parameter count due to extra input features, but the difference is marginal. This setting is simple and maintains comparable expressiveness.

\paragraph{Equal Parameter Count.} For an even stricter comparison, we can adjust the Co-Design network’s hidden layer size downward such that the total number of network parameters matches that of the Control-Only network exactly (eliminating the small increase from the additional inputs). For a single-hidden-layer MLP, one can solve for the adjusted hidden size $H^\sim$ using the parameter count formulas for each network. For example, if the Control-Only network parameter count is:
\[
    P_{\text{ctrl}} \approx (d_{in} \cdot H) + (H \cdot d_{out}) + H + d_{out},
\]
and the Co-Design network (with morphology input of dimension $d_m$) is:
\[
    P_{\text{code}} \approx ((d_{in} + d_m) \cdot H^\sim) + (H^\sim \cdot d_{out}) + H^\sim + d_{out},
\]
we set $P_{\text{code}} = P_{\text{ctrl}}$ and solve for $H^\sim$ (rounding down to an integer). In practice, we found protocol (A) sufficient, but protocol (B) can be used to confirm that any performance gains are not due to a larger parameter count.

\subsection{Simulation Platform}

The experimental setup consists of three components: simulator, optimizer, and controller.  

\textbf{Simulator}: The simulation is based on the MuJoCo physics engine, which provides accurate dynamics, collision handling, and realistic joint behavior. The robot is modeled as a planar 2-DoF RR robotic arm, composed of two revolute joints. To ensure realistic behavior, joint damping and friction are incorporated into the simulation. Target positions for the reaching task are generated within the robot’s initial workspace using a rectangular grid. These target points are confined to the robot’s reachable area, and any target inside an obstacle with a safety margin is excluded, ensuring valid and diverse target configurations.


\textbf{Optimizer}: Morphology and control parameters are evolved using a genetic algorithm (GA). Each individual encodes link lengths and controller weights, with fitness defined as task loss plus collision penalties. The GA runs for a fixed number of generations to ensure fair comparison.  


\textbf{Controller}: A feed-forward neural network maps states to control actions. Both Control-Only and Co-Design conditions employ identical architectures; the only difference is that morphology parameters are included as additional inputs in the Co-Design case, enabling adaptation of both body and controller.

\subsection{Task Definition}


We focus on a static reaching task where the arm’s end-effector must reach a fixed target in the presence of static obstacles, corresponding to the \emph{Level 3 static tasks} illustrated in Fig.~2. To evaluate performance, we adopt a Monte Carlo procedure that samples multiple target positions within the same obstacle environment, assessing how target placement influences reaching accuracy and success.  

Beyond a single layout, we also vary obstacle configurations and repeat Monte Carlo target sampling in each case. This enables a systematic comparison of how morphology--control co-design and control-only methods adapt to different static environments. In all trials, the arm starts from a fixed initial pose and executes until either the target is reached or a time limit is exceeded.

\begin{figure}[htb]
    \centering
    \includegraphics[width=0.8\linewidth]{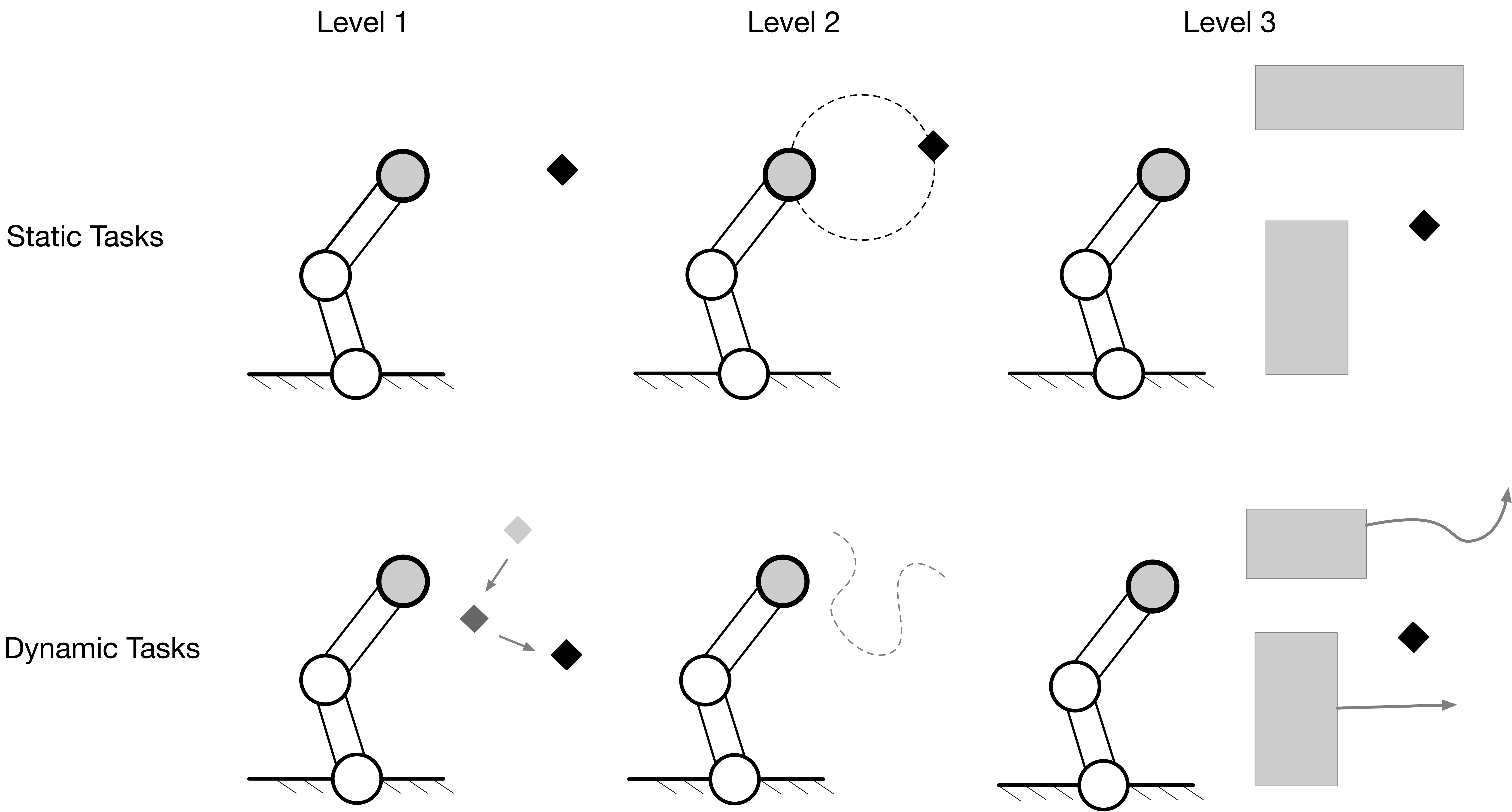}
    \caption{Task hierarchy categorized into static and dynamic tasks with increasing complexity}
    \label{fig:figure1}
\end{figure}

\subsection{Evaluation Metrics}

We assess performance along accuracy, 
safety, and optimization dynamics. Accuracy is quantified by the trajectory error, defined as the mean squared distance between the end-effector and the target over an episode,
\[
\varepsilon_{t} \;=\; \frac{1}{T}\sum_{t=1}^{T} \big\| p_t - x^*_t \big\|_2^2,
\]
where $T$ is the horizon and, for static targets, $g_t \equiv g$. We further report the final error as the terminal Euclidean distance,
\[
\varepsilon_{final} \;=\; \big\| p_T -x^*_T \big\|_2,
\]
which directly measures the proximity to the goal at the end of the rollout. Lower values of TE and FE indicate better tracking and terminal accuracy, respectively.

Task completion is summarized by the success rate over $N$ trials, computed as the average of the indicator of achieving a tolerance $\epsilon$ at the end of the episode:
\[
p_{success} \;=\; \frac{1}{N}\sum_{k=1}^{N} \mathbf{1}\!\left\{ \varepsilon_{final}^{(k)} < \epsilon \right\}.
\]

Safety is incorporated via a cumulative collision penalty that aggregates per-step contact costs,
\[
C_{\mathrm{coll}} \;=\; \sum_{t=1}^{T} c(\mathcal{C}_t),
\]
where $\mathcal{C}_t$ is the set of contact events at time $t$ and $c(\cdot)\!\ge\!0$ depends on, e.g., penetration depth or signed-distance violations. We also report the weighted contribution $\lambda_{\mathrm{coll}}\,C_{\mathrm{coll}}$ used in the optimization objective; lower values indicate safer trajectories.

\section{Results and Case Studies}

\subsection{Single-target case study}
\begin{figure}[htb]
    \centering
    \includegraphics[width=0.8\linewidth]{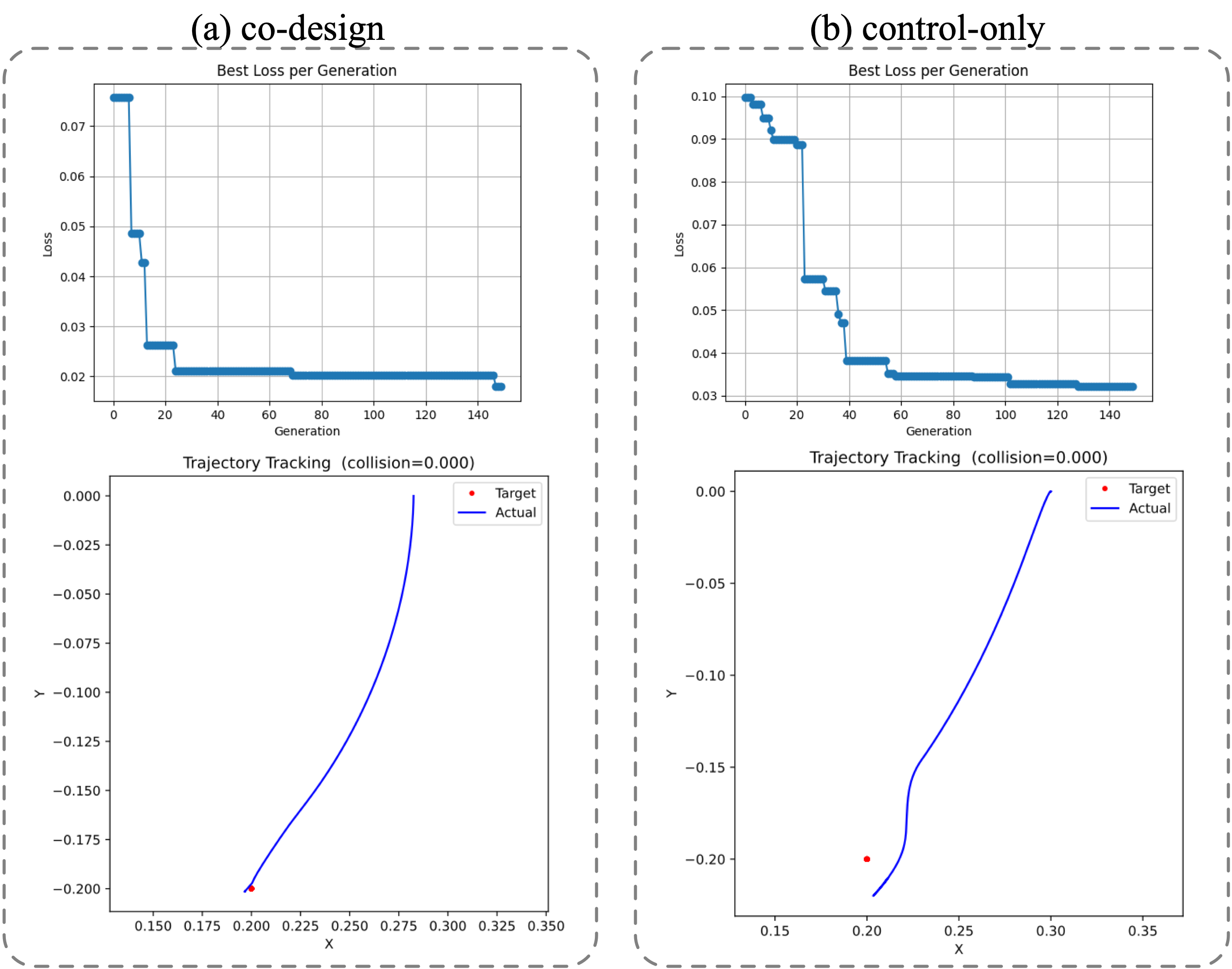}
    \caption{\textbf{Co-design accelerates convergence and reduces tracking error.}
(a) Co-design versus (b) control-only on the same
reaching task with static obstacles, where the top row plots
the best loss per generation and the bottom row shows the end-effector
trajectory (blue) towards the target (red). Loss includes tracking error and a
collision penalty; in this example the reported collisions are zero for both
methods. Co-design converges within $\sim$20--30 generations to a lower
plateau loss ($\approx 0.02$), whereas control-only requires $\sim$100
generations and plateaus higher ($\approx 0.03$), yielding a larger residual
offset in the final trajectory.}
    \label{fig:figure2}
\end{figure}

In Figure~\ref{fig:figure2}, panels (a) co-design and (b) control-only use the identical target and obstacle layout. The top row reports the per-generation best loss and the bottom row shows the
XY-plane end-effector trajectory towards the same target (red dot). In the
\emph{co-design} setting (panel a), the loss drops sharply within $\sim$20--30
generations and reaches a lower plateau ($\approx 0.02$), while the trajectory
exhibits a smooth approach to the target. In contrast, the \emph{control-only}
setting (panel b) reduces the loss more slowly---requiring roughly $\sim$100
generations to plateau at a higher value ($\approx 0.03$)---and the resulting
trajectory shows a larger residual bias and curvature. Both solutions are
collision-free, indicating that the improvement stems from tracking rather than
contact penalties. Overall, the joint optimization of morphology and control
accelerates convergence and achieves lower final error, consistent with the
hypothesis that morphological adaptation simplifies the control problem.

\subsection{Population-level statistics and spatial regularities}

\begin{figure}[!htbp]
    \centering
    \includegraphics[width=0.95\linewidth]{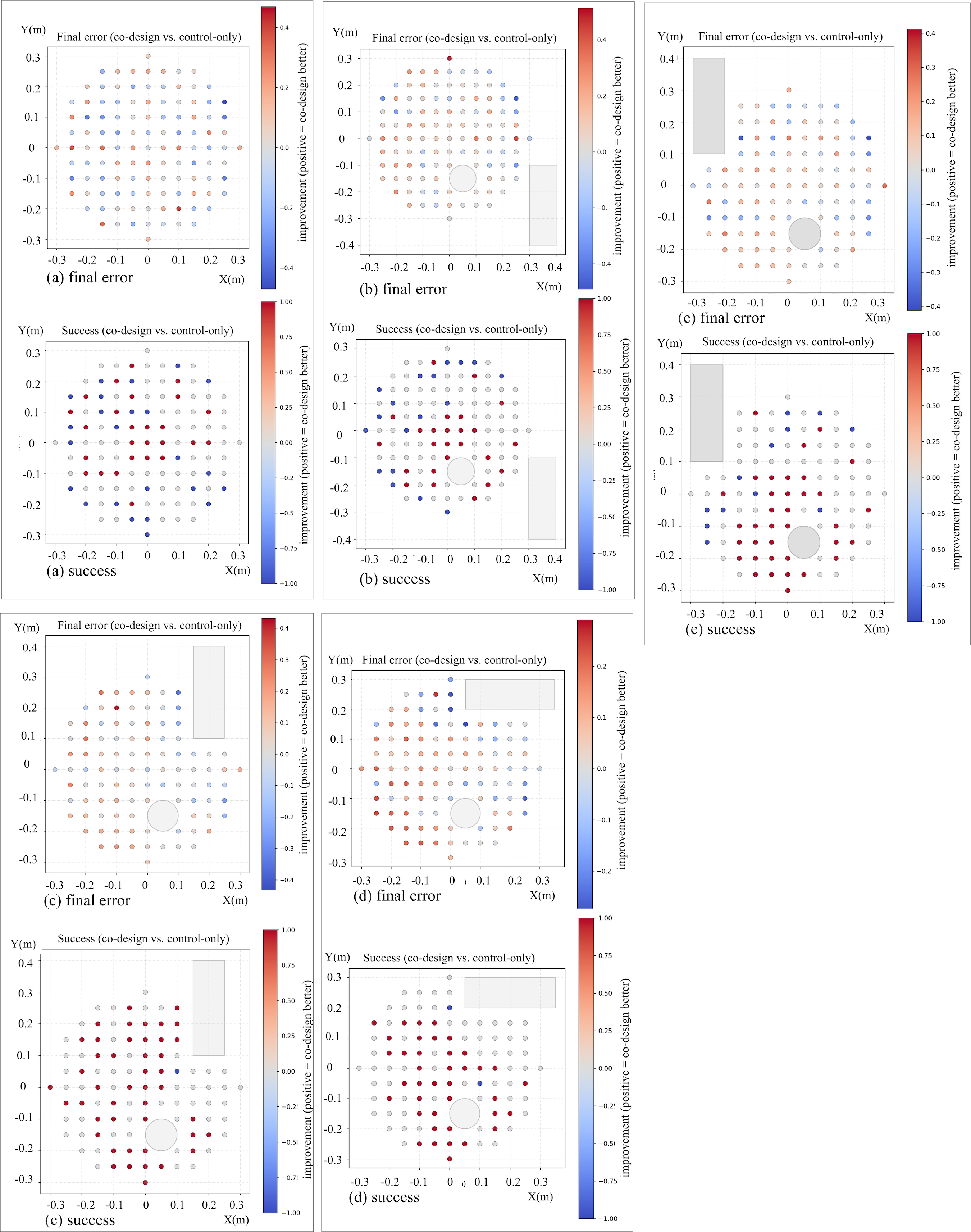}
    \caption{\textbf{patial advantage of co-design across target locations and obstacle layouts.}
Each panel reports a $\Delta$-map over the target grid, with color encoding improvement at each target: 
for error-based metrics we use $\Delta = \text{err}_{\text{control}} - \text{err}_{\text{co}}$, 
and for success we use $\Delta = \text{succ}_{\text{co}} - \text{succ}_{\text{control}}$; thus \textbf{red indicates co-design is better}, and \textbf{blue indicates control-only is better}. }
    \label{fig:figure3}
\end{figure}

According to Figure~\ref{fig:figure3}, co-design yields consistent gains (red) in the vicinity of obstacles and 
near the workspace boundary, where geometric clearance and approach-angle constraints dominate. 
In unobstructed interior regions, the two methods achieve near parity (grey). 
Occasional blue pockets (control-only better) appear at extreme corners or along joint-axis-aligned 
approaches, plausibly reflecting morphology trades (e.g., leverage/inertia sacrificed to enable 
detours) or optimisation variance. These maps indicate that 
co-design should be preferred when collision-avoidance and clearance are critical, whereas control-only 
suffices in open, low-constraint regions.

\begin{figure}[htb]
    \centering
    \includegraphics[width=0.95\linewidth]{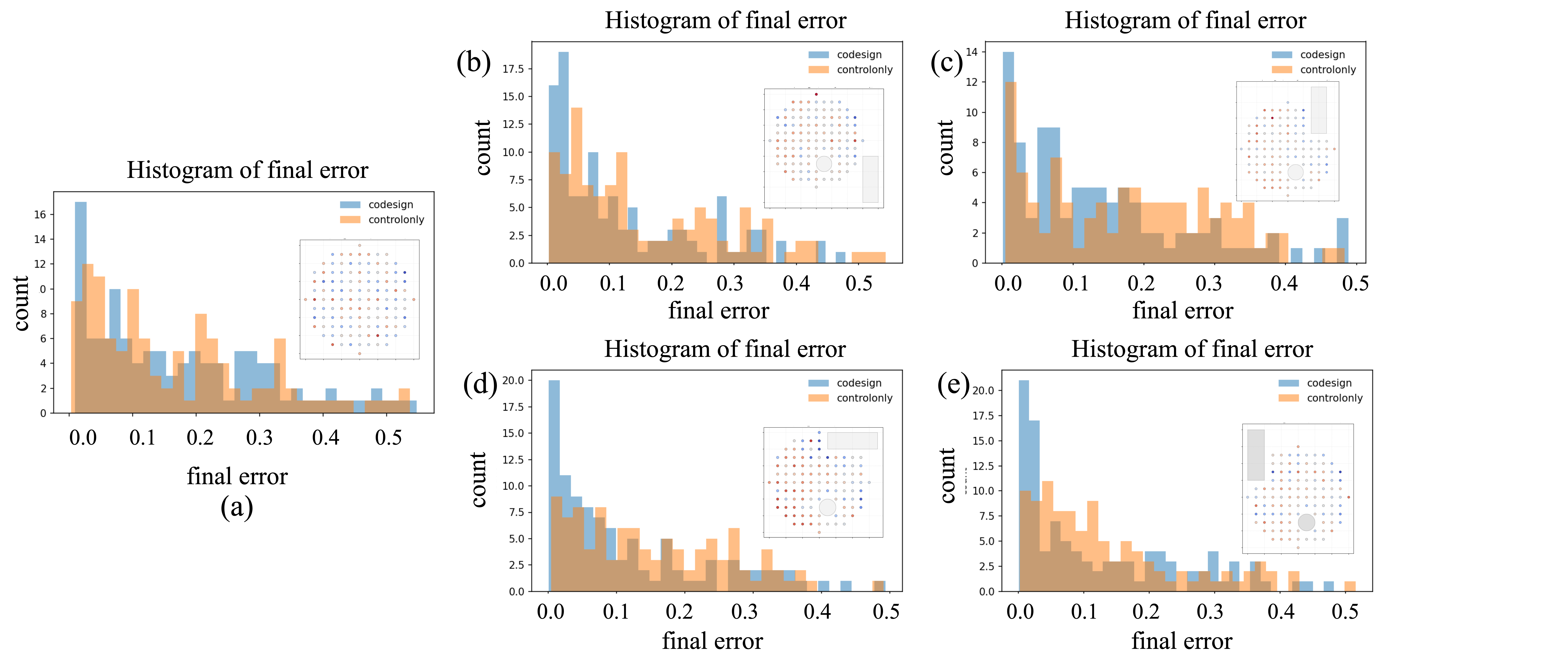}
    \caption{\textbf{Histogram of final position error across obstacle layouts.}
    Each panel (a--e) compares the distribution of end-of-rollout error
    \(\varepsilon_{\text{final}}\) (meters) for \textbf{co-design} (blue) and
    \textbf{control-only} (orange). Insets show the corresponding target grid and
    obstacle geometry (grey). Co-design concentrates more mass in the low-error
    regime (0--0.10\,\text{m}) and suppresses large-error tails (\(\ge\)0.25\,\text{m}) in most
    constrained layouts, whereas distributions largely overlap in open workspaces.
    A localized exception (panel~c) shows a mild control-only advantage around
    0.10--0.20\,\text{m}, consistent with morphology--dynamics trade-offs for clearance.}
    \label{fig:figure4}
\end{figure}

In Figure~\ref{fig:figure4}, panels~(a--e) report histograms of the end--of--rollout position error
($\varepsilon_{final}$, in meters) for \textbf{co-design}
(morphology+control, blue) versus \textbf{control-only} (control with fixed
morphology, orange). Insets depict the corresponding target grid and obstacle
geometry (grey). The $x$--axis is the final error and the $y$--axis is the
count of targets falling into each error bin.

In most layouts---notably (b), (d), and (e)---the co-design distribution
shifts mass toward the \emph{low-error regime} (0--0.10\,m) and suppresses the
\emph{heavy tail} (\(\ge\)0.25\,m), indicating fewer large-error cases when
morphology is co-optimized. Under weak geometric constraints (panel~a), the two
distributions largely overlap, suggesting little headroom for morphology to
help when the workspace is open.

In panel~(c), the \emph{mid-error band} (approximately 0.10--0.20\,m) shows a
slight control-only advantage, plausibly reflecting a morphology--dynamics
trade-off: geometry adapted for clearance may reduce leverage/inertia that
benefits control smoothness, thus yielding less robustness in that band. Notably,
co-design still retains some advantage in the very low-error bins
($\leq 0.03\,m$).

These distributional patterns are consistent with our spatial
\(\Delta\)-maps (co-design minus control-only) and ring/sector groupings:
co-design delivers the largest gains \emph{near obstacles or workspace
boundaries}, while performance parity emerges in interior regions.
Consequently, histograms should be interpreted jointly with the spatial
visualizations to localize where improvements arise.

\begin{figure}[htbp]
    \centering
    \includegraphics[width=0.9\linewidth]{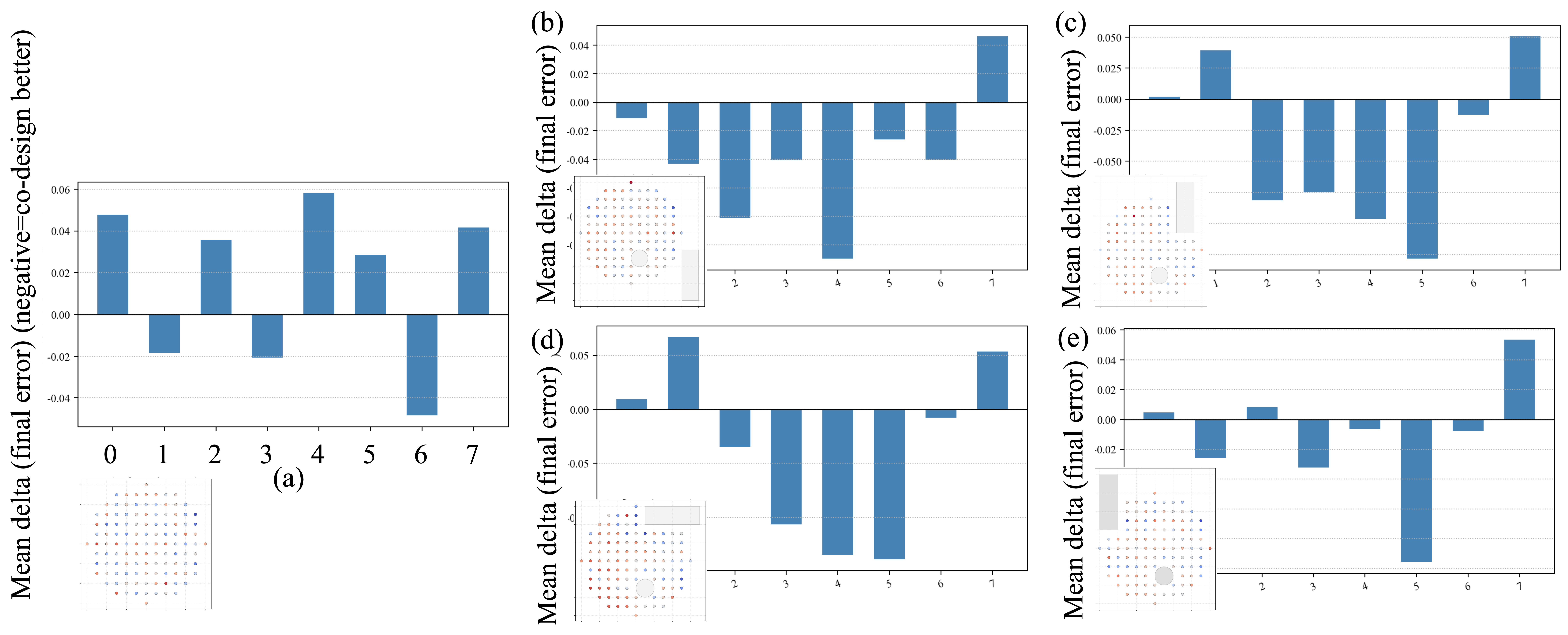}
   \caption{
\textbf{Sector-wise mean improvement in final error across obstacle layouts.}
For each layout (a--e), bars show the sector-aggregated mean
\(\Delta \varepsilon_{final}
=
\overline{\varepsilon_{final}}_{\text{co-design}}
-
\overline{\varepsilon_{final}}_{\text{control-only}}\),
with eight 45\(^\circ\) sectors (0--7) defined around the workspace origin.
\emph{Negative values} indicate that co-design attains lower terminal error
(better performance). Insets depict the target grid and obstacle geometry
(grey). Co-design consistently outperforms in sectors facing obstacles, while
near-open sectors occasionally favour control-only, reflecting trade-offs
between morphology-induced clearance and control robustness.
}
    \label{fig:figure5}
\end{figure}

According to Figure~\ref{fig:figure5}, Panels~(a--e) report bar plots of the mean improvement

\[
\Delta \varepsilon_f \;=\;
\overline{\varepsilon_f}_{\text{co-design}}
-
\overline{\varepsilon_f}_{\text{control-only}},
\]
aggregated over eight 45$^\circ$ angular sectors (0--7) defined around the
workspace origin. The $y$-axis is measured in meters; \emph{negative values
indicate that co-design achieves lower error (better performance)}. Insets show
the corresponding target grid and obstacle geometry (grey).

Across most layouts (b, d, e), sectors that \emph{face the obstacles} exhibit
consistently negative bars, indicating centimeter-level reductions in terminal
error when morphology is co-optimized. This aligns with our $\Delta$-maps and
ring/sector analyses: co-design provides the largest gains in directions where
geometric constraints create kinematic or collision-induced bottlenecks, by
reshaping link lengths, joint limits, and inertial leverage to open smoother
control pathways.

In sectors with \emph{weak geometric constraints} (far from obstacles), small
positive bars occasionally appear, suggesting that when the workspace is open,
holding morphology fixed and optimizing only the controller can be sufficient—%
and sometimes slightly more robust—than introducing morphology-induced dynamic
changes. Hence, sector-wise aggregation complements distributional histograms
and spatial $\Delta$-maps by localizing \emph{where} each method is likely to
dominate.





\subsection{Target-local case studies under static obstacles}
\begin{figure}[htp]
  \centering
  \includegraphics[width=\linewidth]{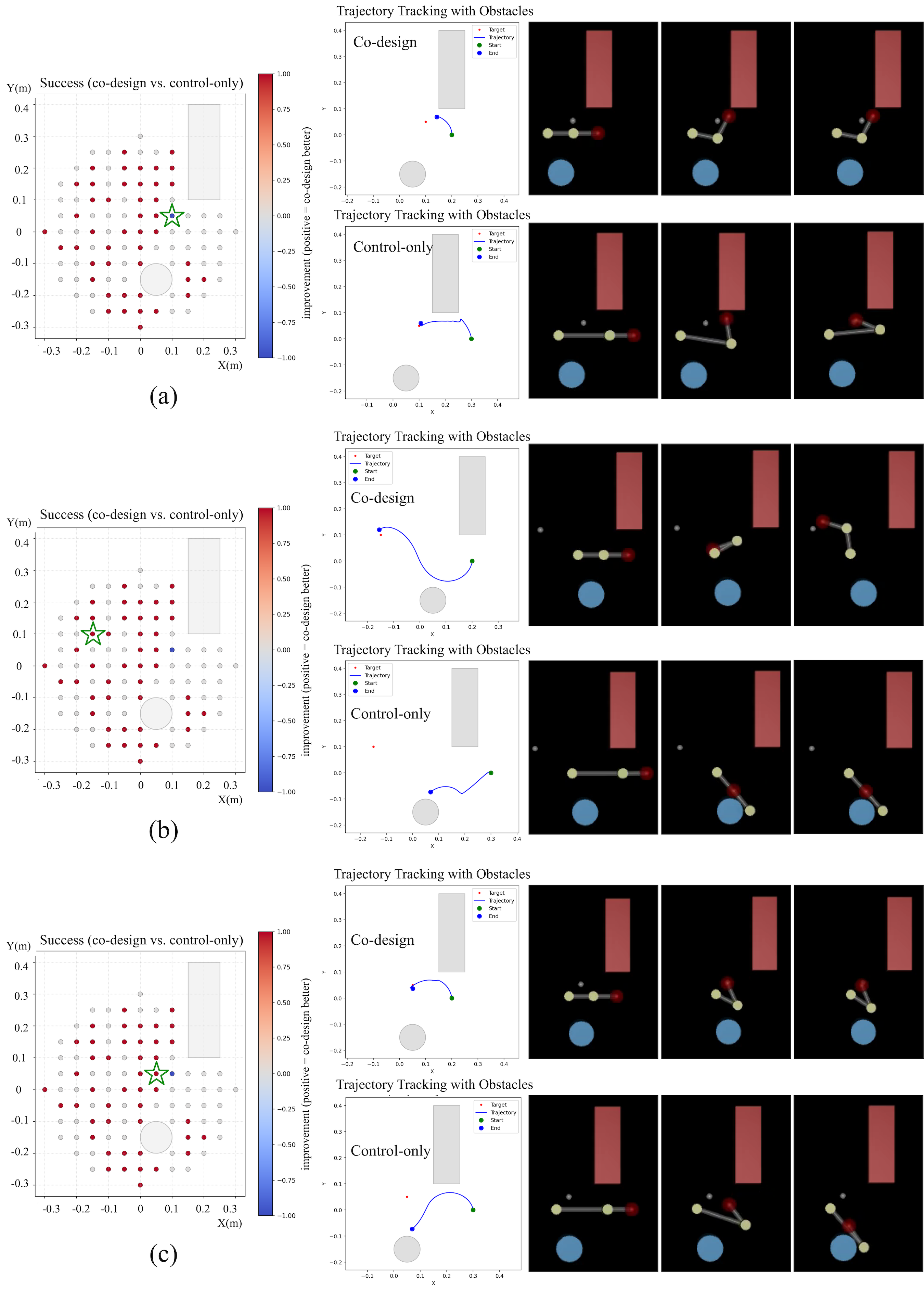}
  \caption{\textbf{Case study 1.} 
  (a) $(0.10,\,0.05)\,\mathrm{m}$ where control-only shows higher success; 
  (b) $(-0.15,\,0.10)\,\mathrm{m}$ where co-design gains advantage due to obstacle clearance; 
  (c) $(0.05,\,0.05)\,\mathrm{m}$ where co-design expands reachability near workspace limits.}
  \label{fig:case1}
\end{figure}

Figure~\ref{fig:case1} illustrates a representative obstacle layout where the relative advantage between control-only and co-design varies across target regions. 

Panel~(a) shows a case where \textbf{control-only achieves higher average success rate}. The target lies in an open sector of the workspace, where the baseline morphology is already sufficient. In this regime, additional morphological adaptation brings little benefit and may even introduce unnecessary structural changes, while control-only converges faster by focusing solely on policy optimization. 

In contrast, Panels~(b) and (c) highlight regions where \textbf{co-design clearly outperforms}. In (b), the target lies close to the obstacle boundary, where avoiding collision requires precise clearance. Here, morphology adapts by adjusting link lengths and angles, effectively opening safer approach corridors that simplify the controller’s burden. In (c), the target is located near the edge of the manipulator’s workspace, partially beyond the baseline reach. Pure control optimization fails due to morphological limits, whereas co-design expands reachability through adjusted link lengths, enabling successful completion. 

Overall, Case~1 demonstrates that \textit{control-only is preferable when the baseline morphology already matches the task geometry}, but \textit{co-design becomes crucial when obstacles constrain feasible paths or when targets approach the limits of the workspace}.

\begin{figure}[htp]
  \centering
  \includegraphics[width=\linewidth]{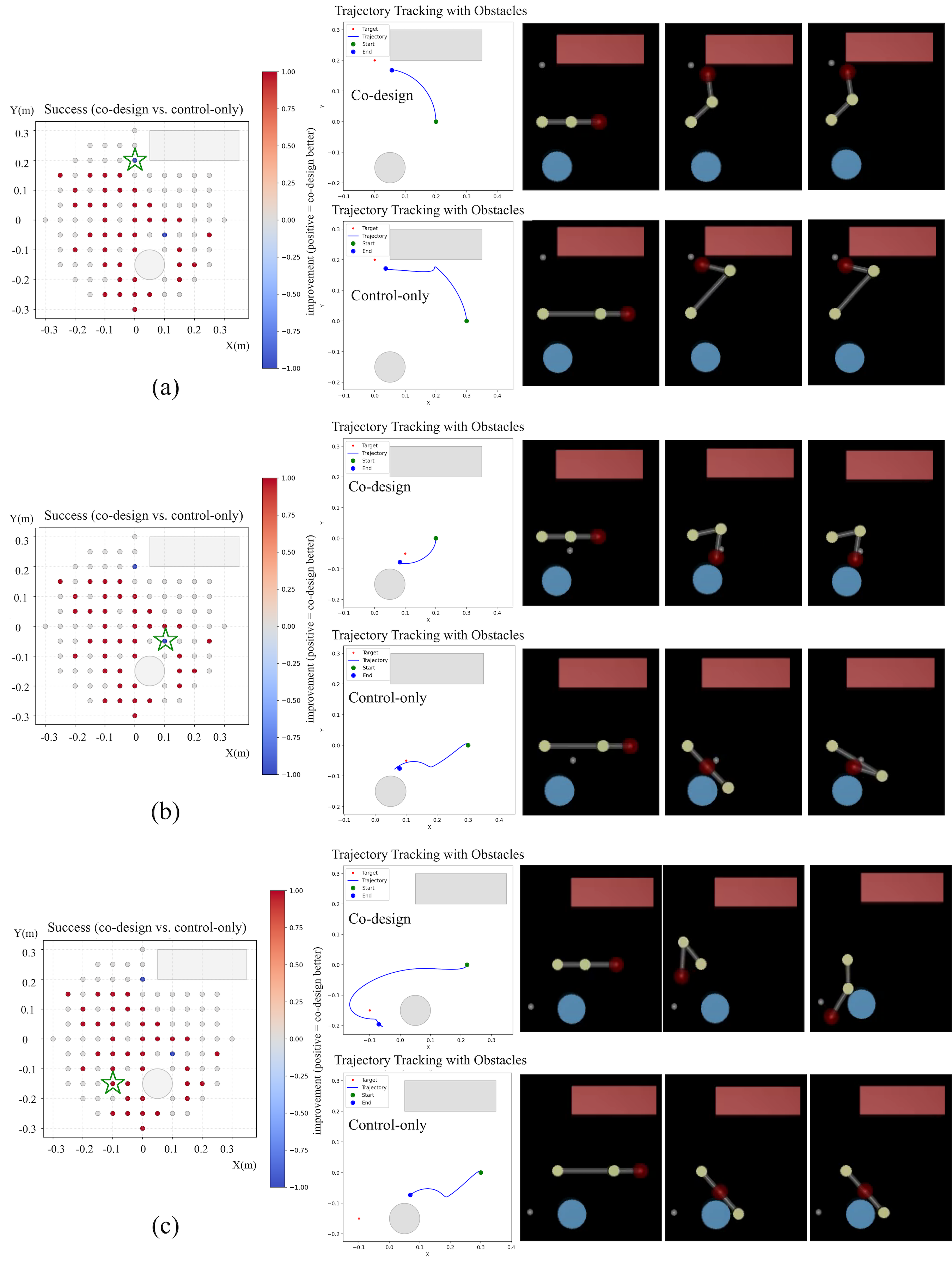}
  \caption{\textbf{Case study 2.}
  (a) $(-0.25,\,-0.05)\,\mathrm{m}$: where control-only performs better, learning a 'hug-the-obstacle' strategy that shortens the approach despite a collision penalty.
  (b) $(-0.05,\,0.25)\,\mathrm{m}$: where control-only again converges faster; co-design’s shape changes add complexity without geometric benefit.
  (c) $(0,\,0.25)\,\mathrm{m}$: wherecontrol-only wins by expanding reach and opening safer approach corridors near the workspace limit.}
  \label{fig:case2}
\end{figure}

In Figure~\ref{fig:case2}, targets lie close to obstacle boundaries. For (a) at $(0.00,0.20)$, 
control-only unexpectedly outperforms co-design: despite collision penalties, the 
policy learns to exploit obstacle margins for efficient access, similar to Case~1(a). 
For (b) at $(0.10,-0.05)$, co-design shows clear gains: morphology adapts link 
lengths to open safer corridors, reducing control burden. For (c) at $(-0.10,-0.15)$, 
control-only again dominates, as the target lies in open space where morphological 
changes add redundancy without benefit. Overall, Case~2 highlights a mixed 
regime: co-design improves near clutter, while control-only remains competitive in open sectors.

\begin{figure}[htp]
  \centering
  \includegraphics[width=\linewidth]{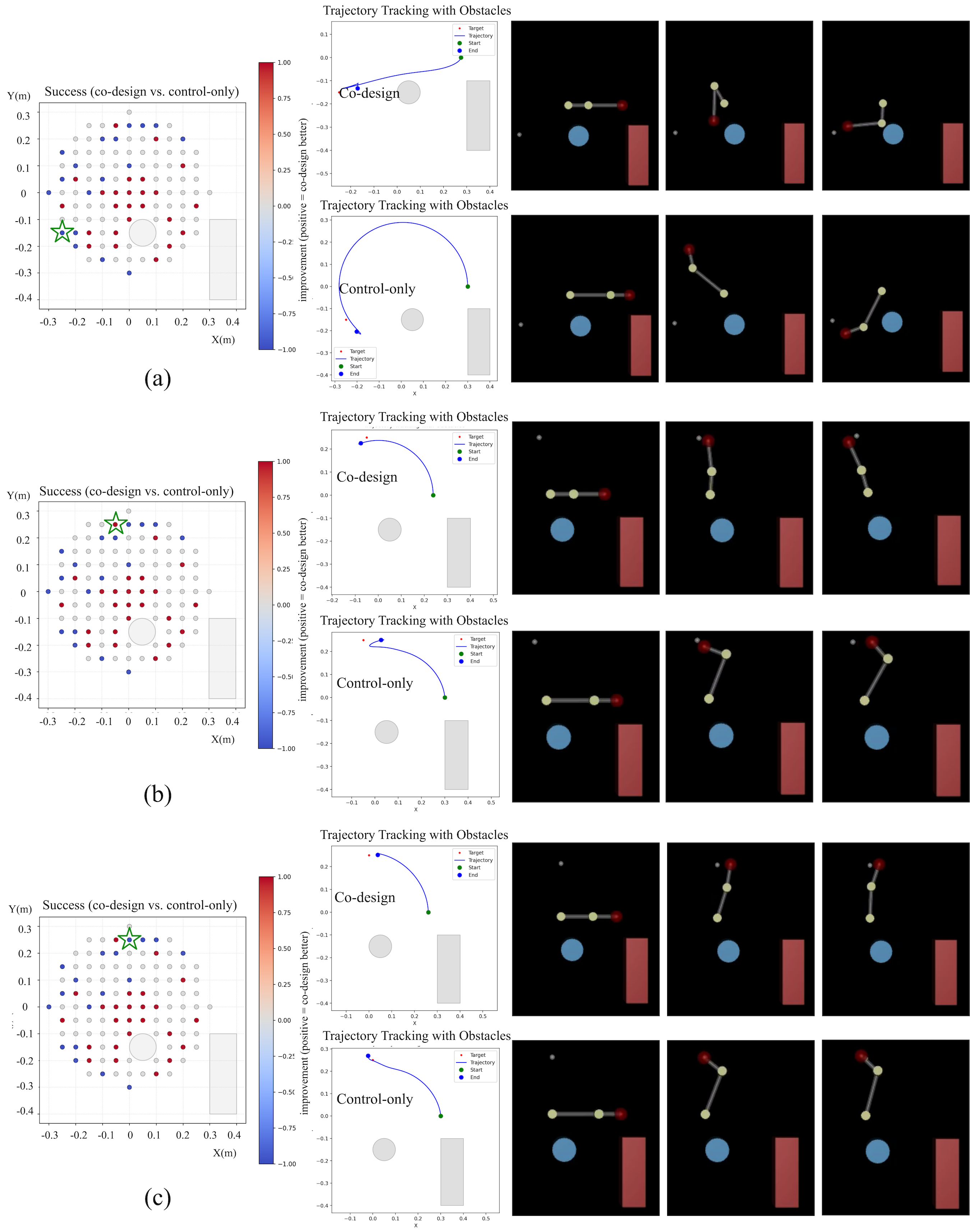}
  \caption{\textbf{Case study 3.} 
  (a) $(-0.25,\,-0.05)\,\mathrm{m}$: control-only outperforms by exploiting inertia for efficient boundary reaching.  
  (b) $(-0.05,\,0.25)\,\mathrm{m}$: co-design prevails by evolving a morphology where the target lies on the new workspace boundary, allowing simpler single-link motion.  
  (c) $(0.00,\,0.25)\,\mathrm{m}$: control-only again achieves better performance, as morphology changes add inertia without structural benefit.}
  \label{fig:case3}
\end{figure}

Case~3(Figure~\ref{fig:case3}) focuses on targets positioned near the boundary of the robot’s initial workspace. 
For (a) at $(-0.25,\,-0.05)$ and (c) at $(0.00,\,0.25)$, control-only achieves higher 
success rates. In these cases, the controller learns more efficient strategies by exploiting 
inertial effects, whereas co-design introduces unnecessary morphological changes that 
increase complexity without improving reach.  

In contrast, (b) at $(-0.05,\,0.25)$ illustrates a clear advantage for co-design. Although 
the target initially lies within the baseline workspace, the evolved morphology shifts the 
workspace such that the target falls on the boundary of the new configuration. This allows 
the robot to reach the target with a simple motion of the first link, significantly reducing 
control effort.  

Taken together, these results suggest that at workspace boundaries, control-only can 
sometimes exploit dynamics to achieve efficient strategies. However, when morphology 
restructures the workspace to align with target locations, co-design provides structural 
advantages by simplifying the required control policy.

\section{Discussion}

\subsection{Why co-design is not always better}
Across obstacle layouts and target grids, co-design does not uniformly dominate control-only.
While co-design increases the hypothesis space by exposing morphological degrees of freedom,
it also enlarges the search space that the optimizer must explore. In our runs, this manifests
as three failure modes: (\textit{i}) slower or stalled convergence for some target sectors
(e.g., sector 7 in Fig.~6 shows a positive $\Delta \varepsilon_{final}$, i.e., control-only wins);
(\textit{ii}) morphology overfitting to local target geometry---the controller compensates less,
but the resulting shape is suboptimal for nearby sectors, leading to heterogeneous $\Delta$-maps;
(\textit{iii}) unstable or inefficient morphologies (e.g., lengthening to reach boundary targets)
that increase inertia and exacerbate overshoot, which the fixed-capacity controller cannot fully cancel.
These patterns are visible in the histograms and ECDFs: the co-design distribution shifts left overall,
yet retains a longer right tail at specific layouts. The per-generation traces confirm that
''more parameters'' can reduce sample efficiency before a good morphology is found.

\subsection{When morphology simplifies control}
Conversely, co-design is consistently advantageous in regions where geometry provides
\emph{structural priors} that simplify the control law. Empirically, the largest improvements occur
(i) near obstacles and narrow passages, and (ii) close to workspace boundaries.
Ring and sector analyses show negative mean $\Delta \varepsilon_{final}$ in inner rings and obstacle-facing sectors (e.g., sector 5 in figure 6),
with win rates $\approx 0.60$--$0.65$ for co-design versus $\approx 0.35$--$0.42$ for control-only.
$\Delta$-maps visualize a coherent spatial pattern: improvements concentrate where kinematic clearance,
manipulability, and collision-avoidance geometry are jointly important.
Single-target case studies corroborate the mechanism: the co-design arm reshapes link length/damping
to maintain clearance and align torque directions with the target manifold,
yielding smoother state-input mappings and requiring less expressive control.

\subsection{Boundary conditions: when to prefer each method}
\paragraph{Co-design is preferable when}
(i) the workspace is constrained by obstacles or tight boundaries,
(ii) controller capacity is limited (small hidden dimension) or actuation is weak,
(iii) the baseline morphology is mismatched to the task (e.g., poor manipulability ellipse at targets),
and (iv) geometric adaptation can open collision-free corridors or reduce nonlinearity of the mapping.
Under these conditions our $\Delta$-maps and grouped success plots show systematic gains.

\paragraph{Control-only is preferable when}
(i) the task is simple and collision-free, (ii) the baseline morphology is already well-tuned,
(iii) the controller has sufficient capacity, and (iv) morphology changes risk degrading global
performance (e.g., boundary sectors where lengthening increases inertia). In such cases,
histograms show smaller errors for control-only and sector-wise $\Delta$ can be positive.

\paragraph{Practical guidance}
To mitigate co-design failure modes, we found the following ablations useful:
(a) morphology trust regions per iteration (limit step size and penalize inertia growth);
(b) multi-target training or target randomization to prevent overfitting to a single sector;
(c) explicit regularizers on collision force, link length sum, and joint damping to keep dynamics stable;
(d) matched controller capacity across methods to ensure fair comparisons; and
(e) reporting paired statistics (per-target $\Delta$) plus ECDFs to reflect heterogeneity rather than means only.
Together, the evidence indicates that morphology \emph{can} simplify control when geometry is the bottleneck,
but not when the controller is already the limiting factor or the baseline shape is well matched.

\section{Conclusion}
This work presents a unified framework for morphology--control co-design, embedding morphology and control parameters within a single neural network for fair, end-to-end optimization. Through case studies in static-obstacle-constrained reaching, we disentangle true morphological benefits from controller capacity effects and reveal clear boundary conditions for the utility of co-design.

Empirically, co-design excels where geometry is the bottleneck: near obstacles and workspace boundaries, or when the controller is capacity-limited. In these cases, the learned morphology opens collision-free corridors, improves manipulability, and smooths the state--input mapping, yielding lower final error and higher success. However, co-design is not uniformly better: in simple, collision-free regions with well-suited baseline morphology and sufficient controller capacity, control-only attains comparable or better accuracy with faster convergence, while unconstrained morphology may overfit or increase inertia.

Our analysis provides actionable guidance: (i) use co-design when tasks demand geometric adaptation or when controller capacity must remain small; (ii) prefer control-only when morphology is already adequate or when adding degrees of freedom would enlarge the search space without structural gains; (iii) analyze paired per-target metrics (ECDFs and $\Delta$-maps) rather than global means. 

Looking ahead, our results suggest the foundation for a benchmark that systematically probes \textbf{when control is enough and when it is not}, extending beyond planar reaching to multi-objective tasks, dynamic targets, and real-world settings. Such a benchmark would enable more rigorous, task-driven evaluation of embodied intelligence and support the principled design of future robots.

\bibliographystyle{unsrt}  
\bibliography{references}  

\begin{thebibliography}{10}

\bibitem{PfeiferBongard2006}
Rolf Pfeifer and Josh Bongard.
\newblock {\em How the Body Shapes the Way We Think: A New View of Intelligence}.
\newblock The MIT Press, 10 2006.

\bibitem{ZieglerNichols1942}
J.~G. Ziegler and N.~B. Nichols.
\newblock Optimum setting for automatic controllers.
\newblock {\em Journal of Dynamic Systems Measurement and Control}, pages 759--768, 1993.

\bibitem{AstromHagglund1995}
Karl~Johan Strm and Tore Hgglund.
\newblock {\em PID controllers: Theory, Design and Tuning}.
\newblock 1995.

\bibitem{GarciaMorari1989}
Carlos~E. García, David~M. Prett, and Manfred Morari.
\newblock Model predictive control: Theory and practice—a survey.
\newblock {\em Automatica}, 25(3):335--348, 1989.

\bibitem{Brockman2016Gym}
Greg Brockman, Vicki Cheung, Ludwig Pettersson, Jonas Schneider, John Schulman, Jie Tang, and Wojciech Zaremba.
\newblock Openai gym, 2016.

\bibitem{Tassa2018DMC}
Yuval Tassa, Yotam Doron, Alistair Muldal, Tom Erez, Yazhe Li, Diego de~Las~Casas, David Budden, Abbas Abdolmaleki, Josh Merel, Andrew Lefrancq, Timothy Lillicrap, and Martin Riedmiller.
\newblock Deepmind control suite, 2018.

\bibitem{Schulman2017PPO}
John Schulman, Filip Wolski, Prafulla Dhariwal, Alec Radford, and Oleg Klimov.
\newblock Proximal policy optimization algorithms.
\newblock volume abs/1707.06347, 2017.

\bibitem{RawlingsMayneDiehl2017}
J.B. Rawlings, D.Q. Mayne, and M.~Diehl.
\newblock {\em Model Predictive Control: Theory, Computation, and Design}.
\newblock Nob Hill Publishing, 2017.

\bibitem{Luo2024TubeMPC}
Yu~Luo, Qie Sima, Tianying Ji, Fuchun Sun, Huaping Liu, and Jianwei Zhang.
\newblock Smooth computation without input delay: Robust tube-based model predictive control for robot manipulator planning.
\newblock In {\em 2024 IEEE International Conference on Robotics and Automation (ICRA)}, pages 10429--10435, 2024.

\bibitem{Sims1994EvolvingCreatures}
Karl Sims.
\newblock Evolving virtual creatures.
\newblock In {\em Proceedings of the 21st Annual Conference on Computer Graphics and Interactive Techniques}, SIGGRAPH '94, page 15–22, New York, NY, USA, 1994. Association for Computing Machinery.

\bibitem{Schaff2019ConstructControl}
Charles Schaff, David Yunis, Ayan Chakrabarti, and Matthew~R. Walter.
\newblock Jointly learning to construct and control agents using deep reinforcement learning, 2018.

\bibitem{Wang2019NGE}
Tingwu Wang, Yuhao Zhou, Sanja Fidler, and Jimmy Ba.
\newblock Neural graph evolution: Towards efficient automatic robot design.
\newblock 06 2019.

\bibitem{Pathak2019SelfAssembling}
Deepak Pathak, Chris Lu, Trevor Darrell, Phillip Isola, and Alexei~A. Efros.
\newblock Learning to control self-assembling morphologies: a study of generalization via modularity.
\newblock Red Hook, NY, USA, 2019. Curran Associates Inc.

\bibitem{Cheney2014SoftRobots}
Nick Cheney, Robert MacCurdy, Jeff Clune, and Hod Lipson.
\newblock Unshackling evolution: evolving soft robots with multiple materials and a powerful generative encoding.
\newblock {\em SIGEVOlution}, 7(1):11–23, August 2014.

\bibitem{RoboGrammar2020}
Allan Zhao, Jie Xu, Mina Konakovi\'{c}-Lukovi\'{c}, Josephine Hughes, Andrew Spielberg, Daniela Rus, and Wojciech Matusik.
\newblock Robogrammar: graph grammar for terrain-optimized robot design.
\newblock volume~39, New York, NY, USA, November 2020. Association for Computing Machinery.

\bibitem{Yi2025SoftGripper}
Sha Yi, Xueqian Bai, Adabhav Singh, Jianglong Ye, Michael~T Tolley, and Xiaolong Wang.
\newblock Co-design of soft gripper with neural physics.
\newblock 2025.

\end{thebibliography}

\end{document}